\documentclass[letterpaper, 10 pt, conference]{ieeeconf}  % Comment this line out if you need a4paper

\IEEEoverridecommandlockouts                              % This command is only needed if 
\usepackage{tabularx}
\usepackage{graphics} % for pdf, bitmapped graphics files
\usepackage{epsfig} % for postscript graphics files
\usepackage{mathptmx} % assumes new font selection scheme installed
\usepackage{times} % assumes new font selection scheme installed
\usepackage{amsmath} % assumes amsmath package installed
\usepackage{amssymb}  % assumes amsmath package installed
\usepackage{comment}
\usepackage{tikz}

\usepackage{booktabs}
\usepackage{xcolor}
\usepackage{bm}
\usepackage{caption}
\usepackage{subfigure}
\usepackage{multicol}
\usepackage{multirow}
\usepackage[pagebackref,breaklinks,colorlinks]{hyperref}
\title{\LARGE \bf
Sparse Points to Dense Clouds: Enhancing 3D Detection with Limited LiDAR Data
}

\author{Aakash Kumar$^{1}$, Chen Chen$^{1}$, Ajmal Mian$^{2}$, Neils Lobo$^{1}$, and Mubarak Shah$^{1}$ % <-this % stops a space
\thanks{$^{1}$Center for Research in Computer Vision, University of Central Florida, USA 
        {\tt\small aakash.kumar@ucf.edu, chen.chen@ucf.edu, niels@cs.ucf.edu, shah@crcv.ucf.edu}}%
\thanks{$^{2}$University of Western Australia
        {\tt\small ajmal.mian@uwa.edu.au}}%
}

\begin{document}

\maketitle
\thispagestyle{empty}
\pagestyle{empty}

%%%%%%%%%%%%%%%%%%%%%%%%%%%%%%%%%%%%%%%%%%%%%%%%%%%%%%%%%%%%%%%%%%%%%%%%%%%%%%%%
\begin{abstract}

3D detection is a critical task that enables machines to identify and locate objects in three-dimensional space. It has a broad range of applications in several fields, including autonomous driving, robotics and augmented reality. Monocular 3D detection is attractive as it requires only a single camera, however, it lacks the accuracy and robustness required for real world applications. High resolution LiDAR on the other hand, can be expensive and lead to interference problems in heavy traffic given their active transmissions.
We propose a balanced approach that combines the advantages of monocular and point cloud-based 3D detection. Our method requires only a small number of 3D points, that can be obtained from a low-cost, low-resolution sensor. Specifically, we use only 512 points, which is just 1\% of a full LiDAR frame in the KITTI dataset. Our method reconstructs a complete 3D point cloud from this limited 3D information combined with a single image. The reconstructed 3D point cloud and corresponding image can be used by any multi-modal off-the-shelf detector for 3D object detection. By using the proposed network architecture with an off-the-shelf multi-modal 3D detector, the accuracy of 3D detection improves by 20\% compared to the state-of-the-art monocular detection methods and 6\% to 9\% compare to the baseline multi-modal methods on KITTI and JackRabbot datasets. Project website link \href{https://bravocharlie-ai.github.io/Sparce_to_Dense/}{https://bravocharlie-ai.github.io/Sparce\_to\_Dense/}

\end{abstract}

%%%%%%%%%%%%%%%%%%%%%%%%%%%%%%%%%%%%%%%%%%%%%%%%%%%%%%%%%%%%%%%%%%%%%%%%%%%%%%%%
%\vspace{-1mm}
\section{INTRODUCTION}
%\vspace{-1mm}
3D detection is essential in robotic vision for applications such as autonomous driving and augmented reality, enabling machines to identify and localize objects in 3D space with sensors like LiDAR. Unlike 2D detection methods, which only provide a 2D bounding box and lack depth information, 3D detection offers details on objects' location, size, and orientation in 3D space. This detailed spatial awareness is crucial for robots to interact safely and effectively with their environments. Consequently, 3D detection techniques are pivotal for advancing machine perception, surpassing the limited accuracy and capabilities of 2D detection methods.

%For autonomous driving, 3D detection can be achieved with two main approaches depending on the sensor types: point cloud-based and multimodal-based. Point cloud-based detection relies on LiDAR data and can be further categorized into voxel-based and point-based methods. Voxel-based \cite{8578570, pointpillar, pvrcnn} methods  involve dividing point clouds into grids and applying 3D convolutions to generate a volumetric representation, which is then fed into a region proposal network. Whereas, point-based methods \cite{9008777, 8954080,9156597, 9157660} use raw point clouds and employ iterative sampling and grouping to create abstract point representations, providing higher accuracy but lower efficiency. Multimodal detection methods \cite{mvxnet, 9578812, yin2021multimodal, yoo20203dcvf, bai2022transfusion}, on the other hand, combine data from multiple sensors, including cameras and LiDAR, to improve the detection accuracy and robustness. Choosing between these methods depends on the specific requirements and limitations of the application, as each approach has its own advantages and disadvantages.
%{\bf Put references to all these methods}

In autonomous driving, 3D detection is primarily achieved through point cloud-based or multimodal-based approaches, each dependent on sensor type. Point cloud-based methods, utilizing LiDAR data, are divided into voxel-based methods \cite{8578570, pointpillar, pvrcnn}, which grid point clouds for volumetric representation to apply  3D convolutions, and point-based methods \cite{ 8954080,9156597, 9157660}, focusing on raw point clouds for iterative sampling and abstract representation, trading off efficiency for accuracy. Multimodal methods \cite{mvxnet, 9578812, yin2021multimodal, bai2022transfusion} integrate multiple sensor data, like cameras and LiDAR, enhancing detection accuracy and robustness. The choice among these methods is dictated by the application's specific needs and constraints.

\begin{figure}[t]
\centering
\includegraphics[width=\columnwidth]{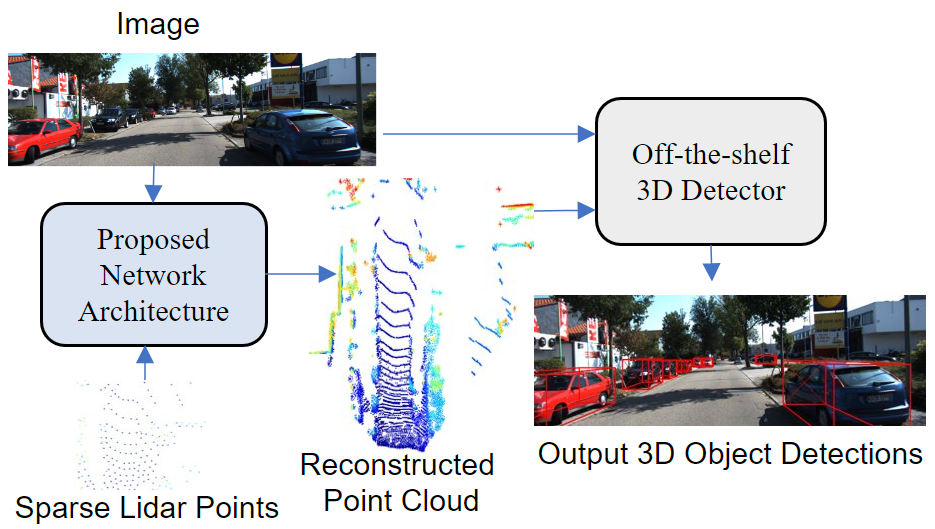}
\vspace{-3mm}
\caption{Overview of the proposed 3D object detection approach. The architecture accepts an input image and a set of sparse LiDAR points, which it then processes to generate a high-resolution point cloud. This dense point cloud, once reconstructed, is paired with the original image and fed into an off-the-shelf 3D object detector, enabling the accurate detection and localization of 3D objects within the scene.}
\vspace{-7mm}
\label{fig:Teaser}
\end{figure}
High-resolution LiDAR and depth sensors have advantages in accuracy and precision, but they are expensive, consume more power and can suffer from interference problems given that they are mostly active sensors. %be vulnerable to weather conditions. 
As a result, researchers have been exploring ways to improve 3D detection performance while reducing costs and complexity. One approach that has gained increasing research attention is monocular 3D detection \cite{MonoDTR, GPUNet, MonoFlex, MonoEF, DFRNet, CaDNN, DDMP}, which estimates the 3D properties of objects in a scene using only a single 2D image.

Monocular 3D detection, while accessible due to its single-camera requirement, faces significant accuracy and reliability issues, making it unsuitable for practical use. Its limitations include poor depth estimation, sensitivity to lighting, and difficulty with complex or occluded shapes. The method also suffers from dependence on precise camera calibration and a restricted field of view. These challenges severely hinder its effectiveness for real-world applications.

To address the aforementioned challenges,  we propose a novel approach to 3D detection by combining monocular vision with a small set of 3D points from a low-resolution and low-cost depth sensor. Our method uses only 512 points or 1\% of a full LiDAR frame of the KITTI dataset \cite{Kitti-dataset} for each detection. Traditional multimodal 3D detection methods relying on high-resolution LiDAR data struggle with lower-resolution LiDAR frame and image data. This is particularly relevant in robotics applications where high-resolution LiDAR is impractical due to power and size constraints. Our method introduces a transformer-based module capable of reconstructing a high-resolution LiDAR frame by processing low-resolution LiDAR inputs accompanied with image data. The reconstructed high-resolution LiDAR data, when combined with the corresponding image, significantly enhances detection precision when processed by standard off-the-shelf multimodal 3D detectors, resulting in a 6-9\% accuracy improvement over baseline methods and over 20\% compared to monocular 3D detection. Overview of our proposed method is shown in figure \ref{fig:Teaser}. This strategy reconstructs full 3D point clouds from few sparse 3D points, improving object detection in autonomous driving and robotics. It seamlessly integrates with existing systems, promising advancements in the field of robotics.

Our major contributions are summarized below:
\begin{itemize}
    \item{We formulate a novel balanced approach between monocular and point cloud-based 3D detection.}
    \item{We propose a novel transformer based architecture for point cloud reconstruction using a single image and a very small number of 3D points.}
    \item{The proposed approach can integrate with 3D sensors of varying resolutions e.g., we show results for 256 to 512 points per scene.}
    \item{We achieve significantly higher mean average precision (mAP) on a popular 3D detection benchmark compared to monocular methods.}
    
\end{itemize}
%Method and contributions
%ur proposed approach has the potential to be applied in various scenarios, where the availability of 3D data is limited, such as in autonomous driving, robotics, and augmented reality. The reconstruction of a complete 3D point cloud from minimal data can help in the efficient processing of data. Furthermore, the use of off-the-shelf detectors for 3D object detection enables the easy integration of other approach into existing systems.

% %\small
% % First version of table.
% \begin{tabular*}{\textwidth}{@{\extracolsep{\fill}} cc|c|cc|cc|cc}
% \toprule
% Query & Neighbouring  & Loss & \multicolumn{2}{c|}{Point Pillar} & \multicolumn{2}{c|}{VoxelRCNN} & \multicolumn{2}{c}{PVRCNN} \\
% Points&Points & Function & Off-the-shelf & Fine-Tuned & Off-the-shelf & Fine-Tuned & Off-the-shelf & Fine-Tuned  \\ 
% \midrule
% 512 & 32 & - & 38.60 & 81.80 & 19.16 & 68.38 & 14.39 & 65.68 \\
% 1024 & 32 & - & 37.59 & 84.59 & 2.32 & 70.67 & 10.24 & 69.14 \\
% 2048 & 32 & - & 38.16 & 86.14 & 2.48 & 50.32 & 11.48 & 54.56 \\
% \midrule
% 512 & 64 & - & 37.77 & 78.21 & 6.94 & 64.23 & 12.89 & 64.78 \\
% 1024 & 64 & - & 44.68 & 88.79 & 2.27 & 74.78 & 12.45 & 73.27 \\
% 2048 & 64 & - & - & - & - & - & - & - \\
% \midrule
% \midrule
% Dummy & 64 & - & - & 65.84 & - & - & - & - \\
% \bottomrule
% \end{tabular*}

\section{Related Work}
\subsection{Monocular 3D Detection}

Monocular 3D detection estimates the 3D features of objects in a scene from a single 2D image, which is challenging due to the lack of depth information. Techniques such as learning 2D features, adding 3D cues, and geometric reasoning are commonly used to enhance the 3D perception from images. Monocular 3D detection finds applications in fields such as autonomous driving, robotics, and augmented reality.
In recent years, there has been an increase in the number of monocular 3D object detection methods that use a single image to estimate the 3D location, dimension and angle of objects \cite{9010867, 9010618, 9327478, 8100080}.  These methods mainly rely on geometric consistency to predict objects since they lack depth information. Deep3Dbox \cite{8100080} uses a unique MultiBin loss to accurately predict the orientation of objects, and enforces a constraint between the 2D and 3D bounding boxes based on geometric principles. MonoPair \cite{9157373} uses a spatial pair-wise relationship among objects to enhance the accuracy of detection. 

Another approach is to learn some depth cues while training the model whereas, inference is still fully monocular. Among these methods, one approach involves combining depth features extracted from the depth map with image features to obtain 2D depth-aware features \cite{D4LCN}. Another approach constructs 3D feature volumes from 2D features to improve the 3D detection capability \cite{CaDNN, imvoxelnet}. Finally, the Pseudo-LiDAR method converts estimated depth maps into Pseudo 3D representations for monocular 3D object detection \cite{ 9009489, 9880232}.

\vspace{-1mm}
\subsection{Pointcloud based 3D Detection}
\vspace{-1mm}

In 3D object detection research, voxel-based and point-based methods stand out. Voxel-based approaches grid point clouds for efficient feature extraction. VoxelNet \cite{8578570} segments scenes into evenly spaced 3D voxels, encodes them with feature layers, and uses 3D convolution to create a volumetric representation for region proposal network processing. SECOND \cite{SECOND} accelerates LiDAR detection training and inference with sparse convolution, introducing angle loss regression. PointPillars \cite{pointpillar} projects 3D clouds onto 2D, using a pillar feature extractor and network for 3D object detection. Point-based methods \cite{ 8954080,9156597, 9157660} focus on raw clouds, employing iterative techniques for precision but at higher computational costs. 3DSSD \cite{9157660} offers an anchor-free, single-stage model with a fusion sampling strategy for efficient downsampling and accuracy. PV-RCNN \cite{pvrcnn} merges 3D voxel CNNs with PointNet abstraction, enhancing feature extraction via voxel set abstraction and RoI-grid pooling for superior 3D proposals and context capture.

\subsection{Mutli-modal 3D Detection}
\vspace{-1mm}
There are different fusion mechanisms for 3D detection that leverage the complementary information provided by LiDAR and camera inputs. Point-level fusion \cite{mvxnet, 9578812, yin2021multimodal} is one approach that combines data at the individual point level, where image features are queried via raw LiDAR points and concatenated back as additional point features. This technique has shown to improve detection accuracy, but it heavily relies on the presence of LiDAR point clouds, which can limit its effectiveness in scenarios where LiDAR input is missing. On the other hand, feature-level \cite{ bai2022transfusion} fusion methods have become the state-of-the-art. However, these methods heavily rely on high resolution LiDAR point clouds.

\section{Methodology}

\begin{figure*}[t]
\centering
\includegraphics[width=\linewidth]{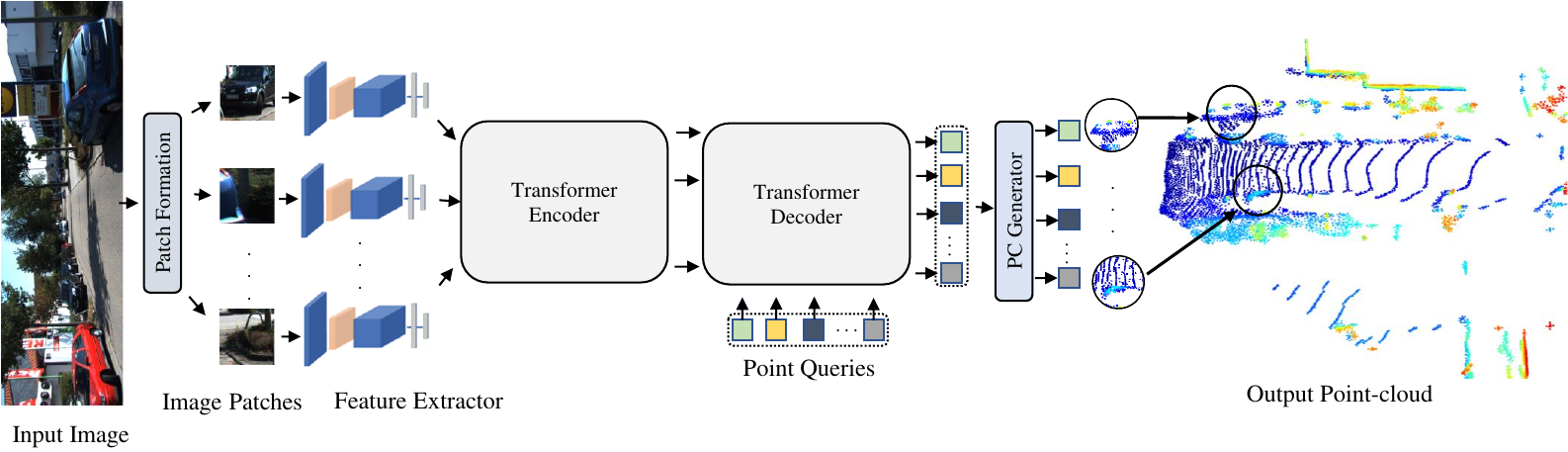}
\vspace{-5mm}
    \caption{Proposed architecture for generating a dense point cloud from an input image and a sparse set of 3D points from a low-cost sensor. Initially, the image is split into 465 patches which are transformed into 256-dimensional vectors by a CNN based feature extractor. These feature vectors combined with sampled 3D points (Point Queries) are passed through a transformer encoder-decoder framework. The encoder uses self-attention to understand patch details, while the decoder employs cross-attention with image-tokens to produce point-tokens for each query point. These tokens are processed by a Point Cloud (PC) Generator that translates them into a dense point cloud. Training involves a Chamfer distance loss function, comparing the predicted point groups with ground-truth data, derived from nearest neighbors to the query points. The outcome is a detailed point cloud useful for 3D object detection and other applications. %, even with low-resolution LiDAR sensor.
% In parallel, each query point is classified as a major or minor point based on the number of points within a 1.2 diameter of the query point. If there are a sufficient number of points within this diameter, the point is classified as a major query point; otherwise, it is classified as a minor query point. As shown in Figure 1, the first and last major queries have neighboring points that are de-normalized for point cloud reconstruction, whereas the two minor queries in the middle are discarded, and their corresponding predictions from the PC layers are also discarded.
}
\label{fig:Architecture}
\end{figure*}

\vspace{-1mm}
\subsection{Problem Formulation}
\vspace{-1mm}
%The network takes two inputs, an image and a set of 3D points, and generates a point cloud as output. 
Given an image and a small set of 3D points as input, the aim is to generate a dense point cloud of the scene corresponding to the image. The input 3D points can be obtained from a low-resolution and low-cost sensor. 
To achieve this, the image is divided into $465$ patches, and each patch is processed through a feature extractor that converts it into a 256-dimensional feature vector. These vectors serve as tokens for the transformer encoder. At the same time, few 3D points are fed into the decoder to perform cross-attention across the image tokens coming from the transformer encoder as shown in figure \ref{fig:Architecture}. This results in a 256-dimensional token for each query point, generated by the transformer decoder. These tokens are then passed through a neural network: the PC generator. The PC generator converts each decoder token into a set of points corresponding to the input queries, allowing for the reconstruction of a dense point cloud corresponding to the input image.

For the input 3D points, we downsample the corresponding LiDAR frame of the image as mentioned in  \ref{Downsampling} to mimic low-resolution and low-cost LiDAR frame. Such frame consists of around $700$ to $800$ points then We do farthest point sampling to select $n$ number of 3D query points $Q_n=\{ q_1, q_2, q_3,...q_n \}$. We show results for $n=$ 256 ,and 512. Against each query point, we predict a group of points $g_i^p=h(q_i)$, where $q_i$ is the $i$-th query point, superscript $p$ stands for prediction and the function $h(.)$ represents the network architecture. Hence, all groups can be represented as $G_n^p = h(Q_n)$. Each $i$-th group of points contains $k$ number of points $ g_i^p = \{ p_{i_1}, p_{i_2}, p_{i_3}...p_{i_k} \}$ where $k$ is 32 or 64. We show results for these settings in the validation table. 

For training the network architecture we need a ground-truth group of points $g_i^g$ against every predicted group of points $g_i^p$. For every query point $q_i$, we select the $k$ neighboring points in a radius of 1.2 meters through random point sampling. These points are then normalize using the input query point, hence, the ground-truth points can be represented as $g_i^g=((p'_{i_1},p'_{i_2},p'_{i_3},...p'_{i_k})-q_i)/1.2$, where the superscript ' represents un-normalized points. All points are normalized with respect to the center query point and diameter. Hence, each normalized set of points range from -1 to 1 $-1<=|p_{i_k}|<=1$. 

%This is how each query point $q_i$ is converted in to group of $k$ points. 
Each query point $q_i$ is converted to a group of $k$ points.  
The output point cloud consists of $n*k$ points. This reconstructed point cloud combined with image can be used for down-streaming tasks such as 3D object detection. Through this method, we can either use low resolution LiDAR, or any other depth sensor that provides some 3D points from the scene to reconstruct the dense point cloud and perform 3D object detection.

% With actual LiDAR query, there are neighboring of points available in the radius of 1.2 meters of radius, hence all groups $G_n^g$ are valid groups. So in this case query classifier is redundant,  Whereas in case of dummy query points, some groups are empty. Figure 2 shows the process of dummy points selection and making the ground truth groups. We select 100 points from each LiDAR frame of training sample through furthest points sampling and combine them, this represent a complete full print of the 3D space where all the object can lie in a scence. Then we again perform furthest point sampling on these points and select 2048 dummy points as show in figure 2. Against every dummy query $q_i$ point than we select neighbouring points $g_i_g$ from the corresponding point cloud of the input image. As shown in figure 2, blue arrow represent a case where neighbouring points are available, whereas red arrow represent a case where points are not available, in this case we repeat the same query $k$ times and this does not contribute in point final point cloud reconstruction. Hence we use query classifier to classify querries as discards queries $Q_D$ and avail queries $Q_A$. As shown in figure 1 , we use only those groups of points which belong to $Q_A$. 
% This simple of dummy point selection can server as a baseline for...

\subsection{Network Architecture}

\subsubsection{Feature Extractor}
Image size in KITTI dataset is $375 \times 1242$, after padding zeros horizontally and vertically, we get a $384 \times 1248$ image that can be divided into equal size patches of $32 \times 32$ resulting in a total of $12 \times 39 = 468$ patches. Padding is performed to avoid changing the 2D scale of objects with respect to their actual 3D scale in the point cloud.  Each patch is passed through VGG-net \cite{vgg-16} with modified skip pooling \cite{Skip_pooling} to pass features from different levels to the top. We employ this feature extractor from \cite{mmmot}. The skip pooling layer is applied at the output of each pooling layer. The number of channels at those pooling layers are 32, 64, 128 and 256. First, global average pooling is applied at each level to gather spatial information and then two point-wise-convolution layers are used to re-scale the number of channels for every layer to 64. Afterwords, these layers are concatenated to form a 256 dimensional feature vector which is fed to the transformer encoder. This type of feature extractor gathers low and high level features from a single patch.

\subsubsection{Image Encoder}
 To capture the global context of the image, we apply self-attention across all the features. This enables the model to learn the global features of the image, facilitating the comparison of objects' appearances with respect to each other and their surroundings, thereby enhancing the ability of the model to discern depth and object size. To achieve this, we use a transformer encoder with four layers of self-attention, each employing multiheaded attention with eight heads. In self-attention mechanism, the input sequence is transformed into three vectors: query, key, and value vectors. The query vector determines the relevance of each part of the sequence to other parts, the key vector identifies which parts the model should pay attention to, and the value vector assigns importance weights to each part.

To compute the relevance of each part, the dot product of the query and key vectors is calculated, and a softmax function is applied to normalize the results into a probability distribution. The value vector is then multiplied by this probability distribution to produce the final weighted sum of the input sequence, where the attention scores determine the weights. Self-attention is useful for learning long-range dependencies and helps the model understand the global context. 

Overall, the combination of CNN-based feature extraction and self-attention mechanism enable the model to learn both local and global features of the image, hence the model can better discern the depth and object size, which is crucial for accurate reconstruction of the point cloud.

\subsubsection{
Decoder}
Transformer decoder takes two inputs, the learned representation from the encoder and the point-query embeddings. The decoder performs both self-attention and cross-attention to generate the output. Cross-attention is performed between each query point and the output tokens of the encoder in the Transformer architecture. This means that the decoder learns to attend to specific regions of the input image corresponding to the query points while generating the output point cloud.

The decoder learns the location of each query point by identifying the relevant parts of the encoder output sequence that are important for that query point. It does this by computing attention scores for each query point based on the dot product between the query vector and key vectors. These scores represent the similarity between the query point and each key vector in the encoder output. The softmax function is then used to normalize the scores to obtain attention weights that are multiplied by the value vectors to obtain the context vector. The context vector represents the weighted sum of the encoder output sequence and captures the relevant information about the image which needed to generate the output point cloud for the corresponding query point. 
%By performing cross-attention for each query point, the decoder learns the location of each query point in the image and the shape of the object at that particular location, which is essential for point cloud reconstruction. 

\begin{figure}[ht]
\centering
\includegraphics[ width=0.9\linewidth]
{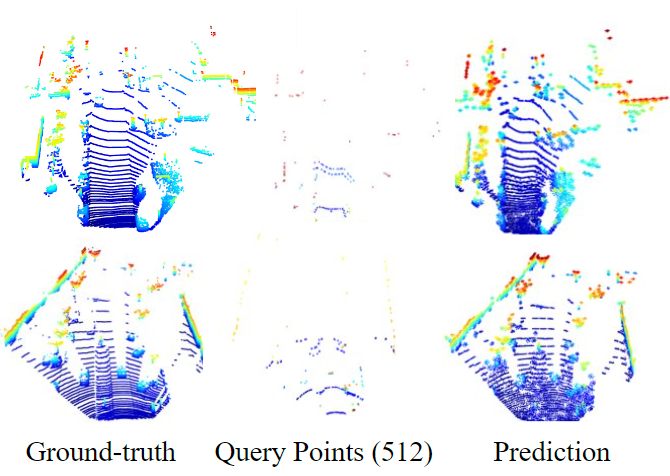}
\vspace{-0mm}
\caption{Ground truth point cloud (LiDAR) compared to point cloud predictions generated using $512$ query points. Each query point generates 32 points. We show the query points with increased point size for better visibility. 
}
\label{fig:q3}
\end{figure}

\subsubsection{
PC Generator}
PC generator consists of four MLP layers. It takes input from the decoder which is of size $k*256$ where $k$ is the number of queries and 256 is the dimension of each token. It converts this 256 dimension vector to a $32*3$ dimension vector representing 32 points, each with 3 dimensions i.e. a point cloud of 32 points. %. Because against every query point we are predicting the 32 points and each point is three dimenstion.

\subsubsection{Query Embeddings.}
We utilize nonparametric embeddings obtained from XYZ locations. Each query point is associated with a query embedding, where the coordinates are converted into Fourier positional embeddings \cite{fourier}.

% First version of table.

\subsection{Loss Function}
We use Chamfer distance loss for pointcloud reconstruction. Chamfer loss measures the distance between the ground-truth and reconstructed point clouds as follows: 
\begin{equation}
d_{\mathrm{chamfer}}(X,Y) = \frac{1}{k} \sum_{x \in X} \min_{y \in Y} \|x-y\|^2 + \frac{1}{k} \sum_{y \in Y} \min_{x \in X} \|x-y\|^2,
\end{equation}
where X and Y are the ground truth and predicted point cloud. In this case $X = g_{i}^{g}$ and $Y = g_{i}^{p}$. Whereas, $x$ is point in sub-point cloud $g_{i}^{g}$ and $y$ is point in sub-point cloud $g_{i}^{p}$. Where $k$ is the total number of points in $X$ or $Y$ which is constant here. We provide results for two settings where $k=32$ and $k=64$. We apply this loss across all $n$ numbers of corresponding groups of prediction and ground truth points.

\begin{table*}[t]
\caption{
Results of monocular 3D detection methods on the KITTI leader board. Our method stands by combining minimal depth information with images to perform 3D detection through point cloud reconstruction. By adding just 512 extra points, our method achieves a significant improvement in performance compared to the monocular and baseline multimodal methods.}
\label{table:leaderboard}
\centering
%\vspace{1mm}
\begin{tabular*}
{0.9\textwidth}{l|c|c|lll|lll}

\toprule
\multirow{2}{*}{\textbf{Method}} & \multirow{2}{*}{\textbf{Reference}} & 
\multirow{2}{*}{\textbf{Modality}} & 
\multicolumn{3}{c}{\textbf{AP\textsubscript{3D}@IoU=0.7}} &
\multicolumn{3}{c}
{\textbf{AP\textsubscript{BEV}@IoU=0.7}} 
\\
% \cmidrule{3-5}
& & & Easy & Medium & Difficult & Easy & Medium & Difficult\\ 
\midrule
\midrule
DDMP \cite{DDMP} & CVPR 2021 & Image & 19.71 & 12.78 & 9.80 & 28.08 & 17.89 & 13.44 \\
CaDNN \cite{CaDNN} & CVPR 2021 & Image & 19.17 & 13.41 & 11.46 & 27.94 & 18.91 & 17.19\\
DFRNet \cite{DFRNet} & ICCV 2021 & Image & 19.40 & 13.63 & 10.35 & 28.17 & 19.17 & 14.84  \\
MonoEF \cite{MonoEF} & CVPR 2021 & Image & 21.29 & 13.87 & 11.71 & 29.03 & 19.70 & 17.26 \\
MonoFlex \cite{MonoFlex} & CVPR 2021 & Image & 19.94 & 13.89 & 12.07 & 28.23 & 19.75 & 16.89 \\
GPUNet \cite{GPUNet} & ICCV 2021 & Image & 20.11 & 14.20 & 11.77 & - & - & -\\
MonoGround \cite{9665899} & CVPR 2022 & Image & 21.37 & 14.36 & 12.62 & 30.07 & 20.47 & 17.74 \\
MonoDTR \cite{MonoDTR} & CVPR 2022 & Image & 21.99 & 15.39 & 12.73 &  28.59 & 20.38 & 17.14 \\
% Ours-512 Points (Pointpillar) & - & I + 1\% Points & 24.80 & 16.87 & 15.70 & 38.52 & 27.24 & 25.82 \\
Pseudo-Stereo \cite{9880232} & CVPR 2022 & Image & 23.61 & 17.03 & 15.16 & 31.83 & 23.39 & 20.57 \\
MonoATT \cite{Zhou_2023_CVPR} & CVPR 2023 & Image & 24.72 & 17.37 & 15.00 & 36.87 & 24.42 & 21.88
\\
MonoDETR \cite{Zhang_2023_ICCV} & ICCV 2023 & Image & 25.00 & 16.47 & 13.58 & 33.60 & 22.11 & 18.60
\\
\midrule
MVX-Net \cite{mvxnet}  & - & Image + 1\% Points & 34.24 & 21.23 & 21.34  & 47.41  & 33.53 & 31.81 \\
Ours $+$ MVX-Net  & - & Image + 1\% Points & 42.61 {\scriptsize(+8.37\%)} & 26.14 & 24.18  & 54.12  & 38.43 & 34.61 \\
\midrule
%Ours-512 Points (MVX-Net)  & old sampling & I + 1\% Points & 48.66 & 29.84 & 25.40  & 63.92  & 43.33 & 37.99 \\
\midrule
EPNET++ \cite{9983516}  & - & Image + 1\% Points & 37.61 & 24.43 & 22.43  & 52.26  & 36.25 & 33.53 \\
Ours $+$ EPNET++  & - & Image + 1\% Points & 44.23 {\scriptsize(+6.62)} & 31.94 & 25.16  & 58.12  & 42.37 & 39.32 \\
\midrule
SFD \cite{Wu_2022_CVPR} & - & Image + 1\% Points & 39.42 & 27.23 & 24.94  & 55.23  & 38.51 & 34.18 \\
Ours $+$ SFD   & - & Image + 1\% Points & 45.13 {\scriptsize(+5.71)} & 31.02 & 27.83  & 59.12  & 43.43 & 41.23 \\
\bottomrule
\end{tabular*}
\vspace{-2mm}
\end{table*}

\subsection{Point-cloud Downsampling}
\label{Downsampling}
In our research, we developed a technique to simulate the data characteristics of a low-resolution LiDAR system, a necessity for one of our key experiments. The initial data was captured using a high-precision LiDAR sensor, featuring 64 lasers/detectors. This sensor covered a vertical field of view spanning $26.8$ degrees in elevation, segmented into $64$ angular subdivisions, each measuring approximately $0.4^{\circ}$ degrees ranging
$+2^{\circ}$ to $-24.8^{\circ}$ from   relative to the horizontal plane. To achieve a reduced vertical resolution, we selectively utilized data from only 8 out of the 64 detectors. This selective sampling effectively narrowed down our focus to $\frac{1}{8}$th  of the original elevation data points, thereby reducing the vertical resolution proportionately. Concurrently, we modified the azimuthal resolution from a high resolution of $0.08^{\circ}$ to coarser $0.64^{\circ}$. This was accomplished by downsampling the original 360-degree azimuthal sweep, selecting every eighth data point from the 4500 high-resolution points (calculated from the $\frac{360^{\circ}}{0.08^{\circ}}$ division), hence sampling at intervals of $0.64^{\circ}$. Additionally, to mimic the performance of a low-cost LiDAR sensor, we introduced an artificial noise level of ±1 cm to the data. This method allowed us to replicate lower-resolution LiDAR data accurately, providing an essential basis for our experimental analysis.

\subsection{Off-the-shelf Detectors}
We use three pre-existing detectors, namely MVX-Net, EPNet++ and SFD. These are multimodal approaches that processes both image and point cloud data.

\subsubsection{MVX-Net}
MVX-Net \cite{mvxnet} improves 3D object detection by combining LiDAR points with image features using two fusion techniques, namely PointFusion and VoxelFusion. MVX-Net extends the VoxelNet algorithm by fusing LiDAR and image data at early stages, and the VFE feature encoding is used to append the pooled image features to every voxel. 
\subsubsection{EPNet++}
EPNet++ \cite{9983516} introduces a Cascade Bi-directional Fusion (CB-Fusion) module and Multi-Modal Consistency (MC) loss for multi-modal 3D object detection, combining hight resolution LiDAR and camera data to enhance feature representation and prediction reliability.
\subsubsection{SFD}
SFD \cite{Wu_2022_CVPR} introduces a multi-modal framework that enhances 3D object detection by generating pseudo point cloud to perform depth completion to address LiDAR data sparsity. It features a novel fusion strategy, 3D-GAF, for integrating features from raw and pseudo point clouds with fine granularity.

\vspace{-3mm}
\section{Experiments}
\vspace{-2mm}
\subsection{Datasets}
\subsubsection{KITTI Dataset}
\vspace{-1mm}
Our proposed approach is evaluated on the KITTI 3D object detection dataset \cite{Kitti-dataset}. The dataset consists of a vast collection of high-resolution images, laser scans, and other sensor data, annotated for various computer vision tasks, including object detection, 3D object detection, and depth estimation. It comprises 7,481 training images and 7,518 test images with corresponding point clouds captured across a midsize city from rural areas and highways.

\subsubsection{JackRabbot Dataset}
The JackRabbot dataset (JRDB) was gathered using a social mobile robot outfitted with a range of sensors, including LiDAR sensor, 360-degree cylindrical stereo RGB cameras, a 360-degree spherical RGB-D camera, as well as GPS and IMU sensors for data collection in diverse indoor and outdoor environments. JRDB offers annotations for pedestrians, featuring 1.8 million 3D oriented bounding box annotations derived from the LiDAR sensors, along with 2.4 million 2D bounding boxes in RGB images from various cameras. Additionally, it includes identity links between the 2D and 3D bounding boxes and across time.

\subsection{Implementation Details}
Our proposed model is implemented using the PyTorch library. We employ the standard \texttt{nn.MultiHeadAttention} module to implement the Transformer model. The encoder consists of 4 layers, and each layer uses multiheaded attention with eight heads. Similarly, the decoder also has 4 layers, closely following the encoder architecture, where each layer uses multiheaded attention with eight heads. To encode the XYZ coordinates in the decoder, we use Fourier positional encodings \cite{fourier}.

We apply a dropout \cite{srivastava2014dropout} of 0.1 for all self-attention and MLP layers in the model. However, we use a higher dropout of 0.3 for the decoder. The model is optimized using the AdamW optimizer \cite{loshchilov2017decoupled}, and a cosine learning rate schedule \cite{loshchilov2016sgdr} is applied to decay the learning rate until it reaches $10^{-6}$. Additionally, we apply a weight decay of 0.1 to prevent overfitting. To prevent the issue of exploding gradients, we apply gradient clipping at an L2 norm of 0.1 \cite{3Detr}.

\subsection{Results and Discussion}

%Monocular 3D detection is a challenging problem that has garnered considerable attention, yet progress has been slow despite numerous proposed methods. In contrast, other approaches such as 3D detection using full/dense point cloud frames or multi-modal fusion have achieved remarkable results. However, these methods require large amounts of data including from a high resolution LiDAR sensor, which may not always be feasible in real-world applications.

%To address this limitation, we propose an 
 Our method requires minimal 3D information and uses  just 512 3D points along with an image for 3D detection. Note that 512 points form only 1\% of the full LiDAR frame in the KITTI dataset. Our approach involves reconstructing a complete 3D point cloud from an image and a limited number of 3D points, followed by the use of off-the-shelf detectors for 3D object detection. This allows us to leverage the advantages of both image and point cloud data while mitigating the challenges of high resolution 3D data acquisition and processing.

\begin{table}[t]
\caption{Results on JackRabbot Dataset}
\label{table:jrdb}
\centering
%\vspace{-1mm}
\begin{tabular*}
{0.5\textwidth}{l|c|cc}

\toprule
\multirow{2}{*}{\textbf{Method}}  & 
\multirow{2}{*}{\textbf{Modality}} & 
\multicolumn{2}{c}{\textbf{AP\textsubscript{3D}}}
\\
% \cmidrule{3-5}
&  & {AP\textsubscript{3D}@IoU=0.3} & {AP\textsubscript{3D}@IoU=0.5} \\ 
\midrule
\midrule
MVX-Net  & I + 1\% Points & 46.52 & 18.65 \\
Ours $+$ MVX-Net   & I + 1\% Points & 52.78 & 23.94  \\
\midrule
%Ours-512 Points (MVX-Net)  & old sampling & I + 1\% Points & 48.66 & 29.84 & 25.40  & 63.92  & 43.33 & 37.99 \\
\midrule
EPNET++  & I + 1\% Points & 48.56 & 27.65 \\
Ours $+$ EPNET++   & I + 1\% Points & 59.54 & 33.54   \\
\midrule
SFD  &  I + 1\% Points & 48.95 & 26.48 \\
Ours $+$ SFD   & I + 1\% Points & 61.48 &  32.48 \\
\bottomrule
\end{tabular*}
\vspace{-6mm}
\end{table}

Figure \ref{fig:q3} gives an insight into the quality of the point clouds generated by our approach. The left column of the figure shows the ground truth point cloud obtained with a LiDAR sensor, and hence considered to be the most accurate 3D representation of the scene. The remaining columns show the input query points (low resolution LiDAR) and the reconstructed point clouds. 

\begin{figure}[b]
\vspace{-4mm}
\centering
\includegraphics[ width=\linewidth]
{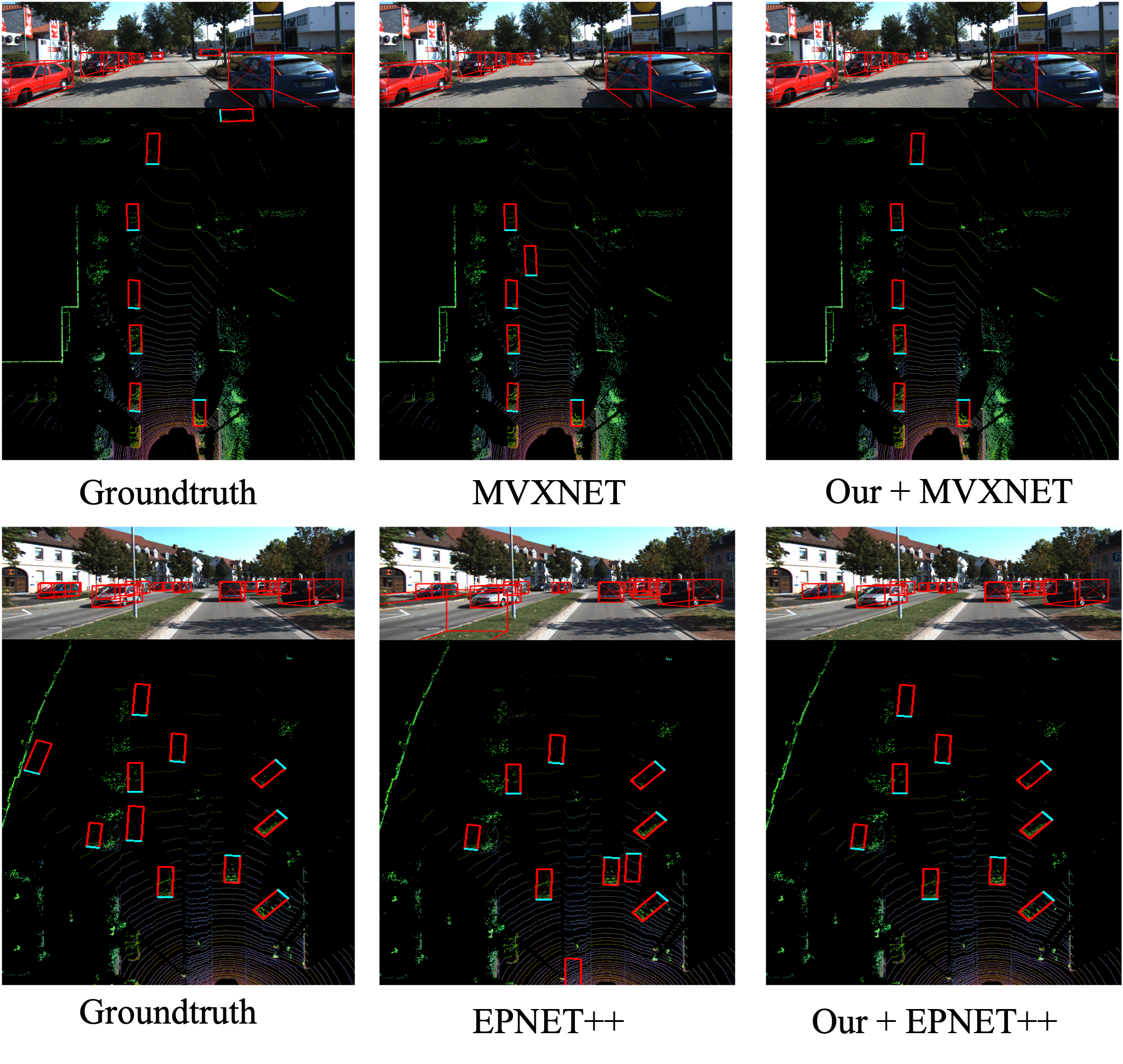}
\vspace{-7mm}
\caption{Qualitative results of 3D detection. 
}
\label{fig:q4}
\end{figure}

\subsubsection{Results on KITTI Test Dataset}
Table~\ref{table:leaderboard}  presents a comparative analysis of various 3D object detection methods on the KITTI test dataset, focusing on their performance across different levels of difficulty and employing distinct modality inputs. Our evaluation is segmented into monocular methods, which rely solely on image data (I), and hybrid methods, which combine image data with a sparse point cloud (I + $1\%$ Points), to demonstrate the impact of incorporating minimal depth information on detection accuracy.
%Monocular methods, including DDMP, CaDNN, DFRNet, MonoEF, MonoFlex, GPUNet, MonoGround, MonoDTR, Pseudo-Stereo, MonoATT, and MonoDETR have been evaluated based on their Average Precision (AP) at an Intersection over Union (IoU) threshold of 0.7. 
Monocular methods, while innovative, illustrate a performance ceiling when tasked with interpreting complex 3D environments from 2D images alone. For instance, MonoDETR, the top performer among monocular approaches, achieves AP scores of 25.00, 16.47, and 13.58 in easy, medium, and difficult categories, respectively, for 3D detection, and 33.60, 22.11, and 18.60 for bird's eye view detection).

To address the limitations of monocular methods, we explored the integration of our novel reconstruction module with off-the-shelf multi-modal detectors (MVX-Net, EPNET++, and SFD), which utilize both image data and a low-resolution point cloud (constituting merely 1\% of a full LiDAR scan). This hybrid approach demonstrates a significant enhancement in detection performance. Notably, the combination of our method with MVX-Net resulted in AP 3D scores of 42.61, 26.14, and 24.18 and BEV scores of 54.12, 38.43, and 34.61 in easy, medium, and difficult categories, respectively, marking a substantial improvement over both the baseline MVX-Net (+8.37\%) scores and the best-performing monocular method, MonoDETR (+17.61\%).

This trend of performance improvement is consistent across all hybrid configurations. For instance, our method combined with EPNET++ and SFD achieved notable gains, surpassing traditional monocular and baseline multi-modal approaches, highlighting the effectiveness of utilizing proposed model for point cloud reconstruction. Moreover, we also show the qualitative results in Figure \ref{fig:q4} which also justifies that by utilizing our proposed architecture detection accuracy is improved. 

\begin{table}[t]
\caption{Ablation study of number of query points(n) and neighbouring points(k). This table shows how varying number of query points and neighboring points affect the detection accuracy. When point cloud reconstructed with different number of query points and neighboring points is given to off-the-shelf-detectors their detection accuracy varies. Best detection accuracy is achieved when we select 512 query points and 32 neighboring points. 
}
\label{table:ablation}
%\vspace{1mm}
\centering
\begin{tabular*}
{0.5\textwidth}{@{\extracolsep{\fill}} cc|c|c|c}
\toprule

%{\textbf{AP\textsubscript{3D}@IoU=0.7}}

Query & Neighboring &
MVX-Net &  EPNet++ & SFD \\
Points & Points & {\text{AP\textsubscript{3D}IoU=0.7}} & \text{AP\textsubscript{3D}IoU=0.7} & \text{AP\textsubscript{3D}IoU=0.7}   \\ 
\midrule
256 & 32 & 63.16 & 67.87 & 69.93\\
512 & 32 & 68.13 & 71.08 & 73.67\\
\midrule
256 & 64 & 62.63 & 66.29 & 67.83\\
512 & 64 & 67.32 & 69.64 & 71.77\\
% \midrule
% \midrule
% Dummy & 64 & - & - & 18.28 & - &  & - & - & - & 50.12\\
\bottomrule
\end{tabular*}
\vspace{-5mm}
\end{table}

\subsubsection{Results on JRDB Test Dataset}
We also provide results on JackRabbot (JRDB) dataset in Table~\ref{table:jrdb} to check the effectiveness of our method on different dataset. We see the similar trend in detection accuracy over three off-the-shelf detectors. Incorporating our module with established off-the-shelf 3D detectors MVX-Net, EPNET++, and SFD demonstrates a significant improvement in 3D object detection on the JackRabbot Dataset, with MVX-Net resulted in an increase in AP from 46.52 to 52.78 at an IoU threshold of 0.3, and from 18.65 to 23.94 at an IoU threshold of 0.5. Similarly, when combined with EPNET++, our module contributed to a substantial increase in AP scores, from 48.56 to 59.54 at IoU=0.3, and from 27.65 to 33.54 at IoU=0.5. SFD witnessed the most pronounced improvement, with AP scores leaping from 48.95 to 61.48 at IoU=0.3, and from 26.48 to 32.48 at IoU=0.5.

The results underscore the potential of our approach in significantly enhancing the accuracy of 3D object detection. By leveraging a minimal set of depth points alongside image data, we demonstrate the feasibility of achieving high-precision detection performance, without incurring the cost and complexity associated with full-resolution LiDAR data. This balanced methodology not only bridges the gap between monocular and LiDAR-based detection systems but also paves the way for more accessible and efficient 3D object detection solutions suitable for a wide range of applications.

\subsubsection{Ablation Studies}
Table \ref{table:ablation} presents an ablation study to evaluate the impact of varying the number of query points and neighboring points on the performance of off-the-shelf 3D detectors, specifically  MVX-Net, EPNet++, and SFD, when operating on reconstructed point clouds generated by our proposed method and combined with image data. The study was conducted using the KITTI validation set, with a focus on understanding how different configurations of query and neighboring points influence detection accuracy as measured by the Average Precision (AP) at an Intersection over Union (IoU) threshold of 0.7.

For the reconstruction of point clouds, our proposed our method takes few number of query points and a corresponding image. For selecting few query points, we employ a downsampling strategy \ref{Downsampling} on the corresponding LiDAR frame associated with an image which result in 700 to 800 points. The we use Farthest Point Sampling (FPS) to select a specific number of 3D points, exploring configurations of 256 and 512 query points. Each query point serves as a seed to predict a group of points, effectively reconstructing the spatial context (neighboring points) around each query location. The number of neighboring points, either 32 or 64, determines the density of the local point cloud around each query point. This setup allows us to simulate various levels of point cloud density and evaluate the robustness of 3D detection methods under different configurations.

\begin{table}[t]
\caption{Ablation study of farthest point sampling and random point sampling. We use our method combined with off-the-shelf detector, MVXNet to check the detection accuracy. Farthest point sampling gives better results.}
\label{table:ablation_fps}
%\vspace{1mm}
\centering
\begin{tabular*}
{0.5\textwidth}{@{\extracolsep{\fill}} cc|c|c}
\toprule
Query & Neighboring &
Farthest Point Sampling & Random point Samping \\
Points & Points & {\text{AP\textsubscript{3D}IoU=0.7}} & {\text{AP\textsubscript{3D}IoU=0.7}}    \\ 
\midrule
256 & 32 & 63.16 & 61.45 \\
512 & 32 & 68.13 & 64.76 \\
\midrule
256 & 64 & 62.63 & 57.74\\
512 & 64 & 67.32 & 63.51 \\
% \midrule
% \midrule
% Dummy & 64 & - & - & 18.28 & - &  & - & - & - & 50.12\\
\bottomrule
\end{tabular*}
\vspace{-5mm}
\end{table}

\begin{table}[b]
\vspace{-5mm}
\caption{This table show the Reconstruction metrics of our method. }
\label{table:reconstruction_metric}
\vspace{-2mm}
\centering
\begin{tabular*}
{0.35\textwidth}{@{\extracolsep{\fill}} ccc}
\toprule
Method &
Chamfer Distance & PSNR  \\
\midrule
Our & 0.058 & 27.54 \\

% \midrule
% \midrule
% Dummy & 64 & - & - & 18.28 & - &  & - & - & - & 50.12\\
\bottomrule
\end{tabular*}
\vspace{-1mm}
\end{table}

The results, as depicted in the Table \ref{table:ablation}, show a clear trend: increasing both the number of query points and the number of neighboring points generally leads to higher AP scores across all three detectors. Specifically, when comparing the configuration of 256 query points with 32 neighbors to 512 query points with the same number of neighbors, a noticeable improvement in detection performance is observed for MVX-Net (from 63.16 to 68.13), EPNet++ (from 67.87 to 71.08), and SFD (from 69.93 to 73.67). This improvement underscores the value of higher query point counts in enhancing the quality of the reconstructed point clouds, thereby facilitating more accurate 3D object detection.

Conversely, the effect of increasing the number of neighboring points from 32 to 64, while keeping the number of query points constant, presents a more nuanced outcome. For the 256 query point setup, the AP scores slightly decrease across all detectors when the number of neighboring points is doubled. This trend is similarly observed for the 512 query point configuration. This indicates that while a denser local reconstruction around each query point might intuitively suggest better performance, the addition of neighboring points beyond a certain threshold may introduce noise or redundancy that does not contribute to, or possibly detracts from, the overall detection accuracy.

These findings highlight the critical balance between the quantity of query points and the density of their associated local reconstructions (neighboring points) in optimizing 3D detection performance. %Employing reconstructed point clouds alongside image data significantly improves the results, demonstrating the effectiveness of our method in leveraging sparse depth information to achieve dense and informative point cloud reconstructions for enhanced 3D object detection. 

We also conduct another ablation study in the Table \ref{table:ablation_fps} to show that using farthest-point-sampling (FPS) on down-sampled point-cloud helps compared to random-point-sampling (RPS). FPS helps because of better coverage of the space, reduced redundancy, and adaptability to different densities. Moreover, to mimic to cost LiDAR frame we add random noise of $\pm$ 1 c.m. To understand the affect of noise level we also perform ablation study on different noise level as shown in table \ref{table:ablation_noise}. Increasing noise level doesn't affect the performance that much because query points servers as a seed to reconstruct neighbouring points around it. Adding different noise levels to query points doesn't affect the overall reconstruction hence detection accuracy is remains intact. We also quantitative result of reconstruction in Table \ref{table:reconstruction_metric}. We achieve chamfer distance of 0.058 and PSNR of 27.54.

\begin{table}[t]
\caption{Ablation study of noise level to mimic the low resolution and low cost LiDAR frame. When different noise level is injected to low resolution LiDAR frame (query point), we see little change in detection accuracy because query points serves just as seed to generate the neighbouring points. This experiment is conduct with n=512 and k=32}
\label{table:ablation_noise}
%\vspace{1mm}
\centering
\begin{tabular*}
{0.45\textwidth}{@{\extracolsep{\fill}} c|c|c|c}
\toprule
Noise Level &
MVX-Net & EPNet++ & SFD \\
centi-meters & {\text{AP\textsubscript{3D}IoU=0.7}} & {\text{AP\textsubscript{3D}IoU=0.7}} & {\text{AP\textsubscript{3D}IoU=0.7}}   \\ 
\midrule
$\pm$0.5 & 67.92 & 70.81 & 73.12\\
$\pm$1 & 68.13 & 71.08 & 73.67\\
$\pm$1.5 & 68.32 & 71.63 & 74.07\\
% \midrule
% \midrule
% Dummy & 64 & - & - & 18.28 & - &  & - & - & - & 50.12\\
\bottomrule
\end{tabular*}
\vspace{-5mm}
\end{table}

\vspace{-1mm}
\section{Conclusion}
\vspace{-1mm}
Monocular 3D detection lacks the accuracy and robustness required for real world applications and high resolution 3D scanners are expensive. We presented a hybrid approach that combines the advantages of monocular and point cloud-based 3D detection by using only a very small number of 3D points. Using as low as 512 points (i.e., 1\% of a full LiDAR frame in the KITTI dataset) and a single image, our method reconstructs a complete 3D point cloud that can be used to improve the performance of existing 3D detectors.

\bibliographystyle{IEEEtran}
\bibliography{egbib}

\end{document}